\documentclass{article}

\usepackage[final,nonatbib]{neurips_2020}

\usepackage[utf8]{inputenc} % allow utf-8 input
\usepackage[T1]{fontenc}    % use 8-bit T1 fonts
\usepackage{hyperref}       % hyperlinks
\usepackage{booktabs}       % professional-quality tables
\usepackage{amsfonts}       % blackboard math symbols
\usepackage{nicefrac}       % compact symbols for 1/2, etc.
\usepackage{microtype}      % microtypography

\usepackage{graphicx}
\usepackage{silence}
\usepackage{menukeys}
\graphicspath{ {./figures/} }

\title{An exploratory experiment on Hindi, Bengali hate-speech detection and transfer learning using neural networks}

\author{%
  Minh Tung Phung \\
  Saarland University\\
  miph00001@stud.uni-saarland.de \\
  \And
  Jan Cloos \\
  Saarland University\\
  s8jacloo@stud.uni-saarland.de \\
}

\begin{document}

\maketitle

\begin{abstract}
  This work presents our approach to train a neural network to detect hate-speech texts in Hindi and Bengali. We also explore how transfer learning can be applied to learning these languages, given that they have the same origin and thus, are similar to some extend. Even though the whole experiment was conducted with low computational power, the obtained result is comparable to the results of other, more expensive, models. Furthermore, since the training data in use is relatively small and the two languages are almost entirely unknown to us, this work can be generalized as an effort to demystify lost or alien languages that no human is capable of understanding. 
\end{abstract}

\section{Introduction}

The emergence of social networks such as Twitter and Facebook inarguably brings up many benefits to humankind. Besides the pure joy of scrolling the new-feeds list, these virtual environments provide people with the easiest way of staying connected to others and also updated to real-time news and social trends. This is especially true during the time of the historic COVID-19 pandemic, when social distancing is more important than ever and one of the most convenient means for people to catch up is to post, comment, and reply on social media. However, some individuals exploit the fact that online platforms are mostly open and anonymous to insult and abuse other people who have different interests, race, color, or religion, etc. A lot of research has been conducted to automatically recognize toxic texts on social media platforms, including work from Gaydhani et al. \cite{gaydhani2018detecting}, which got an impressive accuracy of 95.6\% on classifying English tweets into \emph{Hateful}, \emph{Offensive}, and \emph{Clean}, and from Badjatiya et al. \cite{badjatiya2017deep} with an F1-score of 93\% when categorizing texts into \emph{Racist}, \emph{Sexist}, or \emph{Neither}. While remarkable results were obtained for English, much less work was directed to less popular languages like Hindi. One of such attempts was the organization of the HASOC competition in 2019 \cite{mandl2019overview} in which one sub-task was to process a Hindi dataset (collected from Twitter and Facebook) and differentiate \emph{Hate-speech} from \emph{Non-offensive} content. We also make use of this dataset for our experiment.

One of the common architectures many participants chose was to fine-tune a pre-trained BERT model (Devlin et al. \cite{devlin2018bert}) using the dataset of interest, alongside applying some additional transformations and processing (e.g. Ranasinghe et al. \cite{ranasinghe2019brums}, Mishra and Mishra \cite{mishra20193idiots}). This illustrates how transfer learning has gone viral in the field of natural language processing (NLP). In this paper, we discuss how learning can be transferred from Hindi to Bengali and the benefits of this approach.

\section{Data}
We use the Hindi dataset from the Hate Speech and Offensive Content Identification in Indo-European Languages competition (HASOC) 2019 \cite{mandl2019overview}. The training part consists of 4665 text sentences originated from Twitter and Facebook. Each text has 3 labels corresponding to 3 sub-tasks of the competition: the first label shows whether the text contains any type of offensive content, the second label names the particular type of hate, and the third label determines if the toxic content has specific targets or not. Note that while the vocab size of the corpus is roughly 22K, the total number of tokens is only about 135K, resulting in a low ratio of occurrences per token on average. The test data follows a similar format but contains only 1318 texts. As the goal of the experiment is to distinguish hate and offensive texts from normal contents, only the first label is used.

\begin{figure}[t]
  \centering
  \includegraphics[width=0.9\textwidth]{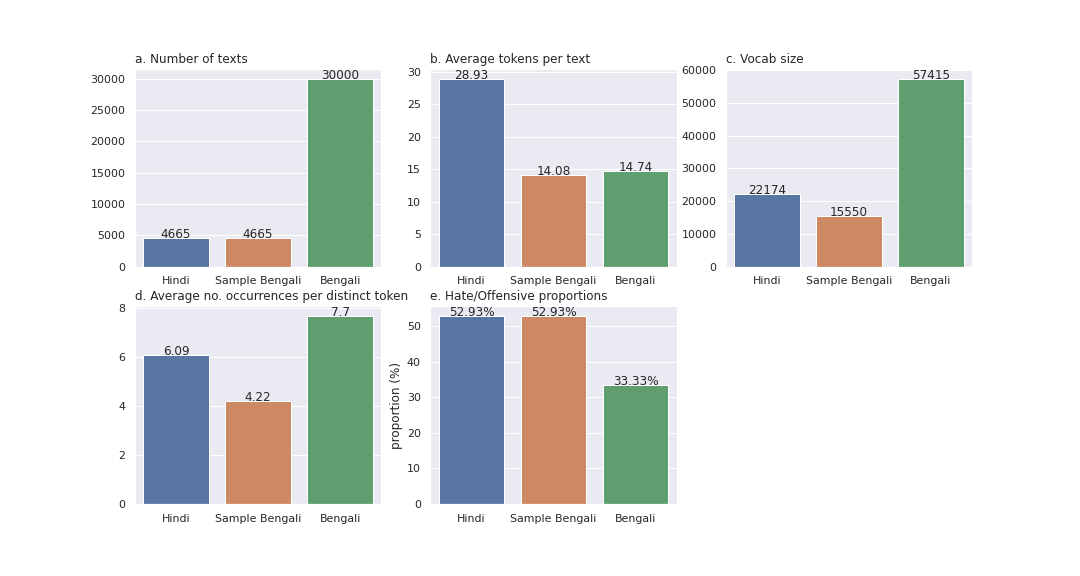}
  \caption{Statistics of the datasets. For the Hindi and sample Bengali, only the training data is considered. Tokenization is done by simply removing punctuations and splitting by whitespace. Stopwords are included in the statistics. a) The number of texts (sentences, comments). b) The average number of tokens per text. c) The vocab size (the number of different tokens). d) The average number of occurrences per distinct token. e) The proportion of hate/offensive texts in the corpus.}
  \label{stats}
\end{figure}

As for the Bengali dataset, we reuse the one from Romim et al. \cite{romim2020hate}. The authors extracted comments from Facebook pages and Youtube videos whose topics range from {\it celebrity} and {\it sports} to {\it crime} and {\it politics}. Each comment is marked with a binary label indicating if it is hate or not. There are 30k labeled comments in total. To alleviate transfer learning and easier comparison with learning-from-scratch, we do random sampling on the Bengali dataset to get a sample of equal size and label distribution as the Hindi dataset for both training and test data. This sample Bengali data is used in the experiments to be described in the next sections. Figure \ref{stats} compares the distributions and statistics of these datasets.

\section{Methodology}

\subsection{Overview}

We break down our work into 3 phases: data pre-processing, training word-embedding, and classification. During the first phase, raw texts are cleaned and tokenized. In the second phase, we pre-train a word-embedding using a word-2-vec model. These embeddings of words help the process of training the text classifier in the third phase more smooth and achieve better performance.

Over the next sections, we present a total of 6 models: a baseline and an optimized model for each of the 3 tasks: training a text classifier of Hindi using Hindi data (Hindi-Hindi), training a text classifier of Bengali using Bengali data (Bengali-Bengali), and training a text classifier of Bengali using Bengali data together with knowledge transferred from the Hindi classifier (Hindi-Bengali).

\subsection{Baseline models}

To create the baseline models for future comparison, we first clean the text by removing the tagging of usernames (those starting with '@'). Since both datasets are originated from comments and posts on social media, a lot of usernames can be found within the texts, however, it is clear that they do not contribute to our objective of hate-speech detection. Punctuations are also gotten rid of with the same argument. Furthermore, we notice that many texts are partially composed of English words. We do not delete but lowercase them for standardization. Stopwords are also dropped as a common practice. Note that we keep the hashtags since they can potentially contain information to link different texts to each other and even more directly, some hashtags might involve explicitly offensive content.

After the cleaning phase, word embedding is trained with the Skip-gram architecture as proposed by Mikolov et al. \cite{mikolov2013efficient}. About the choices of hyperparameters, we use an embedding size of 300 and window size of 10, as suggested in the paper and the corresponding reference code \cite{google2013}. However, even though the authors originally used Adagrad as the optimizer, we replace it with Adam since Adagrad has a drawback about its cumulative penalty that makes the model unable to learn after a certain number of training steps. Adam, on the other hand, has been the default optimizer of most neural networks for several years. The loss function is cross-entropy (the CrossEntropyLoss on Pytorch, which is equivalent to NLLLoss applied on LogSoftmax of the output, so this loss function complies with the original paper).

\begin{figure}[t]
  \centering
  \includegraphics[width=0.8\textwidth]{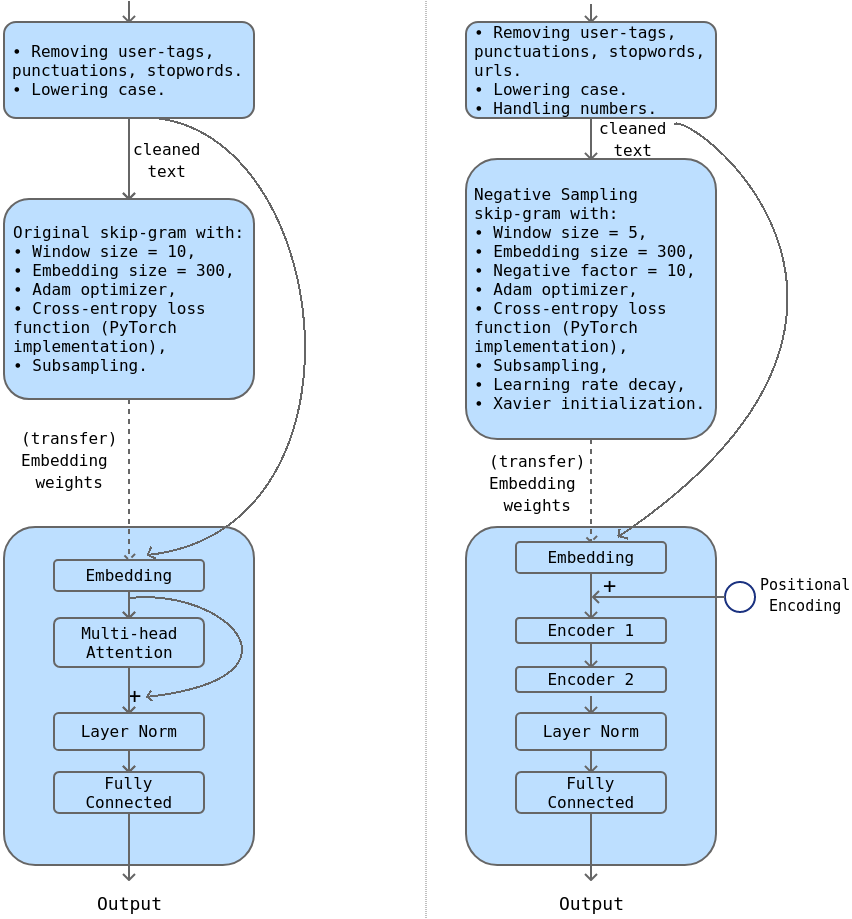}
  \caption{Model architectures: the baseline model pipeline for Hindi-Hindi and Bengali-Bengali text classification is shown on the left, the enhanced model pipeline is shown on the right.}
  \label{architectures}
\end{figure}

The text classifier is a simple neural network with an embedding layer at the bottom (weights of this layer were learned from the previous phase), a multi-head attention layer (MHA) with layer normalization in the middle, and a fully connected at the top to give predictions. There is also a shortcut connection from the embedding layer to the output of the MHA to facilitate training. The binary cross-entropy with logit loss and Adam are chosen as the loss function and the optimizer by default, respectively. Figure \ref{architectures} (left) outlines the three phases of the baseline model.

% While both Hindi-Hindi and Bengali-Bengali models follow the above schedule, the Hindi-Bengali is a bit different. It re-uses the weights of the non-embedding layers from the Hindi-Hindi model. With this, during the training of the classifier, mostly the Bengali word-embedding layer is learned. Since Hindi and Bengali share the same origin, we believe they are similar to a certain extent, which motivates the adoption of transfer learning.

\subsection{Enhanced models}

To improve model performance, we do a number of experiments to better refine data, learn a more meaningful word embedding, and find more suitable network structures. 

We spot that aside from usernames, the text also contains web URLs, which need to be removed. Moreover, there are numbers in the text. Treating each number as a token is probably not effective since most of the numbers only appear once in the corpus, making it hard to learn their suitable embedding. Moreover, numbers should not be entirely deleted as they may enclose useful information. To cope with this, we keep the occurrences of numbers 0, 1, and 2 (since they appear more times and may have specific meaning) but replace all others with some special tokens corresponding to the number of digits. That is, for example, all numbers with 2 digits are replaced with a special token, the numbers with 3 digits are replaced with another special token. This simple replacement (i.e. binning) groups similar numbers into the same bin, whose effect is to increase the ratio of occurrences per token, which in turn facilitates learning. We also try forming 2-gram phrases using the statistical approach as described by Mikolov et al. \cite{mikolov2013distributed}. However, this even reduces model performance. Another attempt was to form 2-grams using only the absolute counts (i.e. without dividing by the counts of 2 individual unigrams), which gives no better results. We suspect the reason is that the small dataset size leads to a non-reliable statistical inference. We will see in the following that other techniques that depend on data statistics also give unfavorable effects on this dataset. One of those attempts was to try using Byte-Pair Encoding (BPE), which was introduced by Sennrich et al. \cite{sennrich2015neural}, with the help of the {\it sentencepiece} library by Kudo \cite{kudo2018}. BPE is a tokenization strategy based on the frequency of adjacent bytes. First, the vocabulary is initialized as the set of all byte-codes in the training corpus. Then, it iteratively computes the most frequent co-occurrences of words to form new symbols to enlarge the vocabulary. This process terminates when the vocabulary size reaches the desired size, which is a hyperparameter the user defined from the start. The BPE tokenizer has no assumptions about the language (some assumptions of other tokenizers are, for example, that words are separated by space, and that dots are used to end a sentence), which makes it seemingly ideal for dealing with entirely unknown languages. However, in our case, using BPE instead of simple splitting by whitespace decreases validation accuracy by about 3 to 4\%.

For the training of word-embedding, we choose the Skip-gram with Negative Sampling, which was proposed by Mikolov et al. \cite{mikolov2013distributed}. According to the authors, this variant of Skip-gram not only gets rid of the expensive cost of the softmax layer in the original Skip-gram model but also achieves very good results in terms of performance. This is achieved by altering the network's objective from predicting nearby words (context) given the input words (center) to assessing if each pair of input words are close to each other. We use the window size value of 5 after trying out a few options. The ratio of negative samples per positive data point is chosen to be 10, this also matches the suggestion in the paper, saying that for small datasets, the value should be in the range 5-20 \cite{mikolov2013distributed}. For noise distribution, we exploit the recommended Unigram distribution raised to the power of $\frac{3}{4}$ (this indeed increases the accuracy of the validation set). Subsampling of frequent words is also kept from the basic models. Moreover, we employ the Xavier initialization for both center and context embedding matrices. The Xavier initialization, introduced by Glorot and Bengio \cite{glorot2010understanding}, allows the networks with {\it tanh} or {\it sigmoid} activation functions to keep the same variance across layers during training, resulting in not only faster but also sometimes better convergence \cite{glorot2010understanding,Ng230}. As we use sigmoid activation in the current network, we try out this initialization method and find that the validation accuracy is increased by about 1 to 3\% for all models. Another alternative to Skip-gram is GloVe \cite{pennington2014glove}. However, as we test this architecture on the Hindi dataset, the classifier's performance decreases by roughly 3 to 4\% in terms of accuracy. Interestingly, Badjatiya et al. \cite{badjatiya2017deep} observed the same negative impact of GloVe on the same task (hate-speech detection) but with a different dataset (which has 16K texts, more than our Hindi dataset but is still quite small). Since GloVe is based on the statistic matrix of global counts over all word co-occurrences, we suppose it is suffering from the same problem we encountered with BPE as described above. Lastly, we apply adaptive learning rate with an exponential decay to balance between fast training speed and good convergence.

About the text classifier, we examine 2 main architectures: Long-Short Term Memory (LSTM) \cite{hochreiter1997long} and the Encoder (from the Transformer by Vaswani et al. \cite{vaswani2017attention}). For LSTM, both the normal (unidirectional) and bidirectional LSTM are studied, with the bidirectional version slightly outperforming the other one. For LSTM models, we experiment with doubling the inputs (e.g. if the original text is "this dish is delicious", we input to the model the doubled text "this dish is delicious this dish is delicious"), inspired by the fact that it is often easier for humans to understand a difficult sentence (or paragraph) by re-reading it one more time. However, the observed changes in accuracy are minimal. We next explore the Encoder architecture and find that both training speed and prediction performance are improved by a large margin. Since the attention mechanism inside the Encoder allows parallel computation, it is expected that the learning progress is faster compared to the sequential process of LSTM. Furthermore, the better performance is also not of a surprise since the Transformers have taken over Recurrent neural networks for the state-of-the-art performance in many different tasks in recent years (e.g. as shown in the work by Lakew et al. \cite{lakew2018comparison} and Karita et al. \cite{karita2019comparative}). 

For the choice of normalization, even though Batch Normalization (Ioffe and Szegedy \cite{ioffe2015batch}) is very popular, it is not straight-forward how it should be applied on text input, where the text lengths (number of tokens) might vary. Instead, we try out Layer Normalization (Lei Ba et al. \cite{ba2016layer}) and Instance Normalization (Ulyanov et al. \cite{ulyanov2016instance}) and find the former one works better and is more stable. This should be due to the fact that Layer Normalization often gives more regularization effects since it is computed over a wider range of the input data. To even enforce more regularization, Dropout is applied after every multi-head attention layer with zeroing-probability of $0.7$. Moreover, we try augmenting the input texts by randomly removing some small portions of tokens and/or replacing them with similar tokens (similarity is computed based on the Cosine-similarity of word-embeddings), but no improvement is recorded. On the other hand, the Sine-Cosine Positional Encoding (as described by Vaswani et al. \cite{vaswani2017attention}) is working well. Since the attention mechanism does not take in the tokens sequentially (as in the case of LSTMs), we need the Positional Encoding to include the ordering information of tokens into the inputs. By adding this encoding to the word-embedding right before feeding it to the first Encoder block, we observe that the maximum validation accuracy rises slightly, and even more importantly, it fluctuates around the highest point as the training loss approaches zero (without Positional Encoding, the validation accuracy may go down by as much as 4-5\% from the highest point as the number of training epochs increases). 

For the Bengali dataset, we take data (excluding the testing set) for learning word-embedding. During the training of the classifier, only the sample training set (which is equal in size to the Hindi training data) is used. 

In the end, we settle down with the overall pipeline as shown in Figure \ref{architectures} (right).

\begin{figure}[t]
  \centering
  \includegraphics[width=0.9\textwidth]{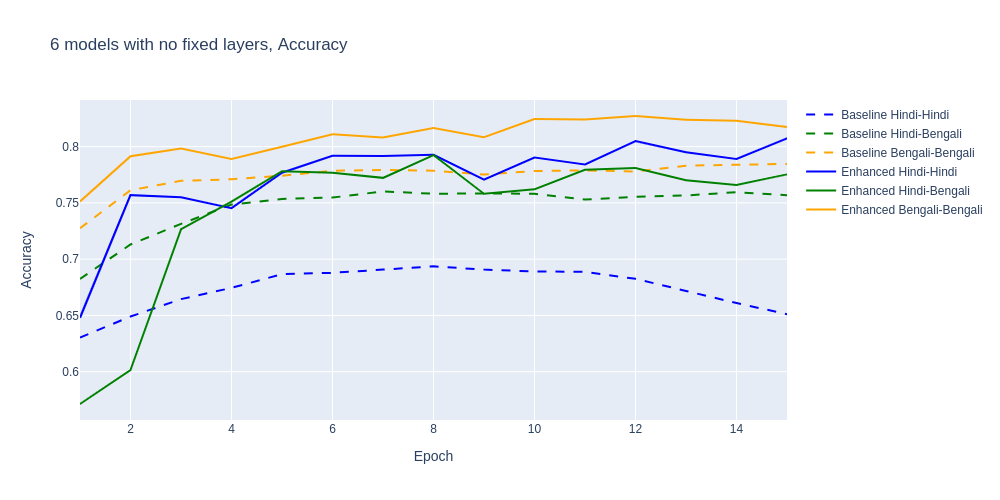}
  \caption{Test accuracy for 3 baseline and 3 enhanced models. No layers are fixed during the last tuning/training phase.}
  \label{accuracy}
\end{figure}

\subsection{Transfer learning}

Since Hindi and Bengali have the same origins in Sanskrit and are both widely spoken in the same region (India), we expect them to share some similarities in grammatical syntax, which motivates transfer learning. In particular, we experiment with reusing the non-embedding layers of the Hindi-Hindi classifier for the Hindi-Bengali classifier. Put differently, the Hindi-Bengali classifier is formed by replacing the word-embedding layer of the Hindi-Hindi classifier with another, untrained word-embedding layer of the Bengali language. During the training of the Hindi-Bengali classifier, either only the word-embedding layer learns (if we fix all the transferred layers) or all layers learn together (if we do not fix any layers, in this case, the transferred weights act as a weight initialization). We explore both cases to see their effects.

With the help of knowledge transfer, we expect the convergent rate of the Hindi-Bengali classifier would be faster than learning from scratch, while still maintain reasonable performance. Furthermore, we hypothesize that if we fix all the transferred weights (i.e. the non-embedding layers), the resulting Bengali word-embedding may likely be forced to converge to an equivalent representation to the Hindi word-embedding of the Hindi-Hindi classifier. In other words, after training the Hindi-Bengali classifier, we may be able to obtain a rough translation between Hindi and Bengali languages using the two word-embedding matrices.

\section{Results}

\begin{figure}[t]
  \centering
  \includegraphics[width=0.9\textwidth]{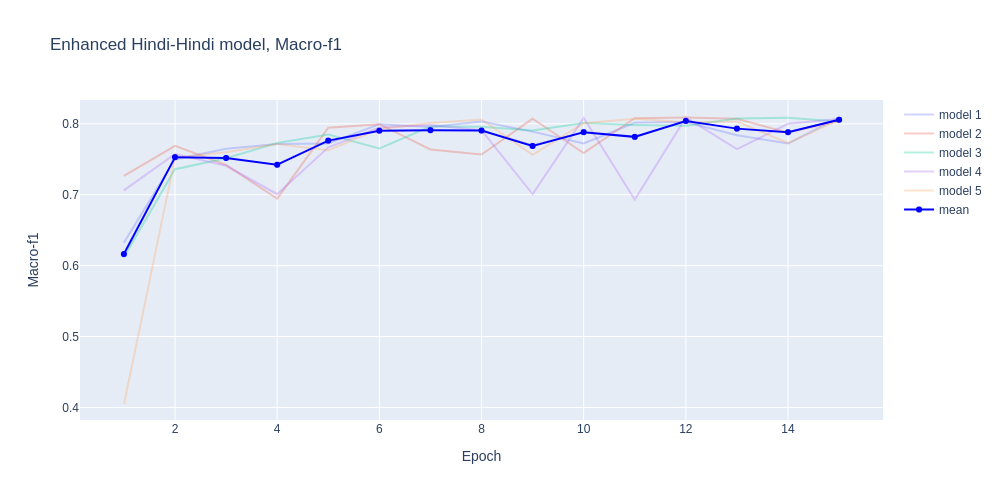}
  \caption{The Macro-f1 score of the enhanced Hindi-Hindi model, taking over 5 different runs.}
  \label{macro}
\end{figure}

To tune the hyperparameters, we first split the data into a big part (80\%) and a small part (20\%). The models are then trained on 80\% of the data and the hyperparameters are adjusted based on their performance on the left-out 20\%. When hyperparameter tuning is done, we retrain the models once again on all training data and evaluate them using the test set. The results are reported based on taking the average of 5 runs.

First, we examine how the models behave if no layers are fixed during the training of the classifier. In other words, we use transfer learning as a weight initialization method. The result is shown in Figure \ref{accuracy} We observe that all enhanced models outperform their baseline counterparts. The biggest rise comes from the Hindi-Hindi models, in which a lift of about 10\% in accuracy is recorded. Moreover, while the baseline classifier tends to get worse after reaching the highest accuracy after epoch 8, the enhanced version does not suffer from the same problem. Regarding the Bengali-Bengali models, the average gain is around 4\%, noticing the fact that the baseline's performance is already high (the average accuracy from epoch 10 onward is 78.2\%). The knowledge-transferred models, Hindi-Bengali, show a slight increment of about 2\%.

It is interesting to see that the performance-decaying effect only happens to the (baseline) Hindi-Hindi but not the Bengali-Bengali model, even though the two datasets share the same size and label distribution. The source of the problem may lie in the differences in vocabulary size, the average text size, the number of occurrences per distinct token, or the intrinsic properties of these languages themselves. This remains an unanswered question. Nevertheless, none of the more-complicated, enhanced models suffers from the same issue.

\begin{table}[t]
  \caption{Best accuracy over all tasks}
  \label{best-accuracy}
  \centering
  \begin{tabular}{llll}
    \toprule
                & Hindi-Hindi       & Hindi-Bengali     & Bengali-Bengali \\
    \midrule
    Config      & enhanced, no-fix  & enhanced, fix     & enhanced, fix \\
                &                   & non-embedding     & embedding \\
    Accuracy    & 80.73\%           & 80.40\%           & 82.98\% 
    \vspace{0.015 \textheight}
  \end{tabular}
\end{table}

\begin{figure}[t]
  \centering
  \includegraphics[width=0.9\textwidth]{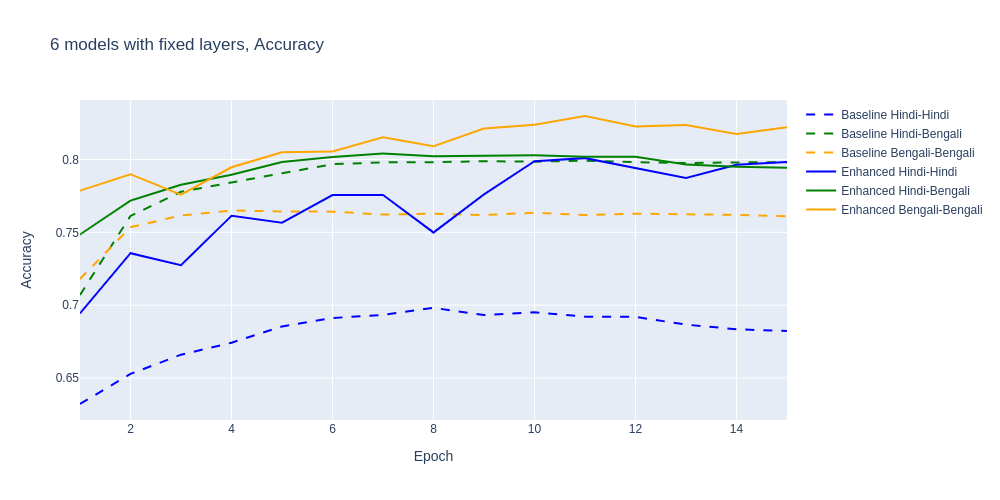}
  \caption{Test accuracy for 3 baseline and 3 enhanced models. The embedding layer of the Hindi-Hindi and Bengali-Bengali models, and non-embedding layers of the Hindi-Bengali models are kept unchanged during the training of the classifiers.}
  \label{accuracy2}
\end{figure}

Figure \ref{macro} gives more insight into the performance of the enhanced Hindi-Hindi model. It shows the Macro-f1 score of all 5 component runs together with the average. After 15 epochs, the average score reaches 0.8057. This result, if to be compared with the teams competed in the HASOC 2019 competition \cite{mandl2019overview}, can be well placed in the top 10.

Next, we inspect the models in case the transferred knowledge (for Hindi-Hindi and Bengali-Bengali models, the embedding weights are transferred from the Skip-grams, and for Hindi-Bengali models, the non-embedding weights are reused from the Hindi-Hindi classifiers) is fixed during the training of the classifiers. This might be a more commonly-used practice of transfer learning in general. The result is shown in Figure \ref{accuracy2}. While the other 2 models still exhibit significant advancement over their baselines, there is no big change in the accuracy of the enhanced Hindi-Bengali model. In fact, this is mostly caused by the surprisingly good performance of the Hindi-Bengali baseline, it gives approximately 80\% accuracy on the test set from epoch 6 onward.

The best test accuracy over all configurations are shown in Table \ref{best-accuracy}.

\begin{figure}[t]
  \centering
  \includegraphics[width=0.7\textwidth]{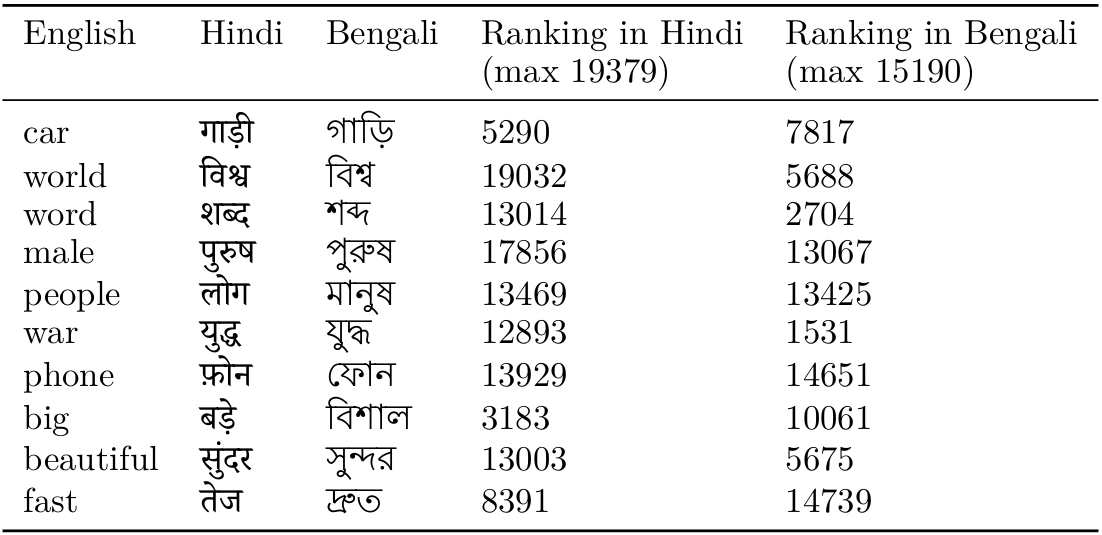}
  \caption{Hindi and Bengali word-to-word translation using Cosine-similarity ranking. \\A ranking of 1 means perfect translation.}
  \label{translation-table}
\end{figure}

Lastly, we qualitatively examine if the word-to-word translation can be exploited as a side effect of transfer learning. We pick 10 common-sense words in English. For each of these, we get its equivalent translations in Hindi and Bengali using Google Translate, and compute the Cosine-similarity ranking of the 2 embedding vectors in comparison to other words in the 2 vocabularies. We use the embedding weights of the enhanced Hindi-Hindi and Hindi-Bengali models for this test. The Hindi-Bengali classifier is trained with all non-embedding layers transferred (from the Hindi-Hindi classifier) and fixed. Figure \ref{translation-table} shows the result. Unfortunately, no sight of meaningful translation can be observed. We hypothesize that the Bengali embedding layer has failed to converge to an equivalent representation as of the Hindi embedding. This might be due to the small datasets, sub-optimal optimizations, and/or the differences between the 2 languages. More work needs to be conducted to investigate this issue.

\section{Conclusion}

In this paper, we explore how to build a hate-speech detection model pipeline either from scratch or through transfer learning. Despite the simple architecture and low computational cost, the models give comparable performance to more complex models like BERT. With transfer learning, we show that by reusing the Hindi model and tuning only the word-embedding layer, we can obtain a Bengali model with marginal difference in performance with a model built from scratch. This shows a sign of simpler transfer learning compared to common practice when we often need to tune several fully connected layers of the pre-trained models to match our tasks. We also make the first attempt towards deriving a word-to-word translator from transfer learning. This could serve as a good direction for future research. 

\bibliographystyle{abbrv}
% Bibliography written in report.bib to cite papers by \cite{key}
\bibliography{report} 

\end{document}